\documentclass[11pt,a4paper]{article}
\usepackage[hyperref]{acl2021}
\usepackage{times}
\usepackage{latexsym}
\usepackage{amsfonts} 
\usepackage{graphicx} 

\usepackage{hyperref}
\usepackage[noend]{algpseudocode}
\usepackage{algorithm}

\newcommand{\pushcode}[1][1]{\hskip#1\dimexpr\algorithmicindent\relax}
\newcommand*\Let[2]{\State #1 $\gets$ #2}

\usepackage{microtype}

\aclfinalcopy 
\def\aclpaperid{2912} 

\title{TAN-NTM: Topic Attention Networks for Neural Topic Modeling}

\author{Madhur Panwar$^{2}$\thanks{\ \ equal contribution}\ \thanks{\ \ work done during summer internship at Adobe} , Shashank Shailabh$^{3}$\footnotemark[1]\ \footnotemark[2] , Milan Aggarwal$^{1}$\footnotemark[1] , Balaji Krishnamurthy$^{1}$ \\ Media and Data Science Research Labs, Adobe$^{1}$ \\ Birla Institute of Technology and Science, Pilani (BITS Pilani), India$^{2}$ \\ Indian Institute of Technology Kanpur (IIT Kanpur), India$^{3}$ \\ \href{mailto:mdrpanwar@gmail.com}{\nolinkurl{mdrpanwar@gmail.com}}, \href{mailto:shailabhshashank@gmail.com}{\nolinkurl{shailabhshashank@gmail.com}}}

\date{}

\begin{document}
\maketitle
\begin{abstract}
Topic models have been widely used to learn text representations and gain insight into document corpora. To perform topic discovery, most existing neural models either take document bag-of-words (BoW) or sequence of tokens as input followed by variational inference and BoW reconstruction to learn topic-word distribution. However, leveraging topic-word distribution for learning better features during document encoding has not been explored much. To this end, we develop a framework TAN-NTM, which processes document as a sequence of tokens through a LSTM whose contextual outputs are attended in a topic-aware manner. We propose a novel attention mechanism which factors in topic-word distribution to enable the model to attend on relevant words that convey topic related cues. The output of topic attention module is then used to carry out variational inference. We perform extensive ablations and experiments resulting in $\sim 9$ - $15$ percentage improvement over score of existing SOTA topic models in NPMI coherence on several benchmark datasets - 20Newsgroups, Yelp Review Polarity and AGNews. Further, we show that our method learns better latent document-topic features compared to existing topic models through improvement on two downstream tasks: document classification and topic guided keyphrase generation. 
\end{abstract}

\section{Introduction}


Topic models \cite{steyvers2007probabilistic} have been popularly used to extract abstract topics which occur commonly across documents in a corpus. Each topic is interpreted as a group of semantically coherent words that represent a common concept. In addition to gaining insights from unstructured texts, topic models have been used in several tasks of practical importance such as learning text representations for document classification \cite{nan-etal-2019-topic}, keyphrase extraction \cite{wang-etal-2019-topic}, understanding reviews for e-commerce recommendations \cite{jin-etal-2018-combining}, semantic similarity detection between texts \cite{peinelt-etal-2020-tbert} etc.


Early works on topic discovery include statistical methods such as Latent Semantic Analysis \cite{doi:10.1002/(SICI)1097-4571(199009)41:6<391::AID-ASI1>3.0.CO;2-9}, Latent Dirichlet Allocation (LDA) \cite{blei2003latent} which approximates each topic as a probability distribution over word vocabulary (known as topic-word distribution) and performs approximate inference over document-topic and topic-word distributions through Variational Bayes. This was followed by Markov Chain Monte Carlo (MCMC) \cite{andrieu2003introduction} based inference algorithm - Collapsed Gibbs sampling \cite{griffiths2004finding}. These methods require an expensive iterative inference step which has to be performed for each document. This was circumvented through introduction of deep neural networks and Variational Autoencoders (VAE) \cite{kingma2013auto}, where variational inference can be performed in single forward pass.




Neural variational inference topic models \cite{miao17a, ding-etal-2018-coherence, srivastava2017autoencoding} commonly convert a document to Bag-of-Words (BoW) determined on the basis of frequency count of each vocabulary token in the document. The BoW input is processed through an MLP followed by variational inference which samples a latent document-topic vector. A decoder network then reconstructs original BoW using latent document-topic vector through topic-word distribution (TWD). VAE based neural topic models can be categorised on the basis of prior enforced on latent document-topic distribution. Methods such as NVDM \cite{pmlr-v48-miao16}, NTM-R \cite{ding-etal-2018-coherence}, NVDM-GSM \cite{miao17a} use the Gaussian prior. NVLDA and ProdLDA \cite{srivastava2017autoencoding} use approximation to the Dirichlet prior which enables model to capture the fact that a document stems from a sparse set of topics.








However, improving document encoding in topic models in order to capture document distribution and semantics better has not been explored much. In this work, we build upon VAE based topic model and propose a novel framework \textbf{TAN-NTM:} \textit{Topic Attention Networks for Neural Topic Modeling} which process the sequence of tokens in input document through an LSTM \cite{10.1162/neco.1997.9.8.1735} whose contextual outputs are attended using Topic-Word Distribution (TWD). We hypothesise that TWD (being learned by the model) can be factored in the attention mechanism \cite{bahdanau2014neural} to enable the model to attend on the tokens which convey topic related information and cues. We perform separate attention for each topic using its corresponding word probability distribution and obtain the topic-wise context vectors. The learned word embeddings and TWD are used to devise a mechanism to determine topic weights representing the proportion of each topic in the document. The topic weights are used to aggregate topic-wise context vectors. The composed context vector is then used to perform variational inference followed by the BoW decoding. We perform extensive ablations to compare TAN-NTM variants and different ways of composing the topic-wise context vectors.\par
For evaluation, we compute commonly used NPMI coherence \cite{aletras2013evaluating} which measures the extent to which most probable words in a topic are semantically related to each other. We compare our TAN-NTM model with several state-of-the-art topic models (statistical \cite{blei2003latent, griffiths2004finding}, neural VAE \cite{srivastava2017autoencoding, wu-etal-2020-neural} and non-variational inference based neural model \cite{nan-etal-2019-topic}) outperforming them on three benchmark datasets of varying scale and complexity: 20Newsgroups (20NG) \cite{lang1995newsweeder}, Yelp Review Polarity and AGNews \cite{zhang2015character}. We verify that our model learns better document feature representations and latent document-topic vectors by achieving a higher document classification accuracy over the baseline topic models. Further, topic models have previously been used to improve supervised keyphrase generation \cite{wang-etal-2019-topic}. We show that TAN-NTM can be adapted to modify topic assisted keyphrase generation achieving SOTA performance on StackExchange and Weibo datasets. Our contributions can be summarised as:

\begin{itemize}
    \item We propose a document encoding framework for topic modeling which leverages the topic-word distribution to perform attention effectively in a topic aware manner.
    \item Our proposed model achieves better NPMI coherence ($\sim$9-15 percentage improvement over the scores of existing best topic models) on various benchmark datasets. 
    \item We show that the topic guided attention results in better latent document-topic features achieving a higher document classification accuracy than the baseline topic models.
    \item We show that our topic model encoder can be adapted to improve the topic guided supervised keyphrase generation achieving improved performance on this task.
\end{itemize}

\section{Related Work}


Development of neural networks has paved path for Variational Autoencoders (VAE) \cite{kingma2013auto} which enables performing Variational Inference (VI) efficiently. The VAE-based topic models use a prior distribution to approximate the posterior for latent document-topic space and compute the Evidence Lower Bound (ELBO) using the reparametrization trick. Since our work is based on variational inference, we use ProdLDA and NVLDA \cite{srivastava2017autoencoding} as baselines for comparison. The Dirichlet distribution has been commonly considered as a suitable prior on the latent document-topic space since it captures the property that a document belongs to a sparse subset of topics. However, in order to enforce the Dirichlet prior, VAE methods have to resort to approximations of the Dirichlet distribution.

Several works have proposed solutions to impose the Dirichlet prior effectively. \citet{Rezaee2020ADV} enforces Dirichlet prior using VI without reparametrization trick through word-level topic assignments. Some works address the sparsity-smoothness trade-off in dirichlet distribution by factoring dirichlet parameter vector as a product of two vectors \cite{Burkhardt2019DecouplingSA}. Wasserstein Autoencoders (WAE) \cite{tolstikhin2017wasserstein} have led to the development of non-variational inference based topic model: Wasserstein-LDA (W-LDA) which minimizes the wasserstein distance, a type of Optimal Transport (OT) distance, by leveraging distribution matching to the Dirichlet prior. We compare our work with W-LDA as a baseline. 
\citet{zhao2021neural} proposed an OT based topic model which directly calculates topic-word distribution without a decoder.

Adversarial Topic Model (ATM) \cite{wang2019atm} was proposed based on GAN (Generative Adversarial Network) \cite{NIPS2014_5423} but it cannot infer document-topic distribution. A major advantage of W-LDA over ATM is distribution matching in document-topic space. Bidirectional Adversarial Topic model (BAT) \cite{wang-etal-2020-neural} employs a bilateral transformation between document-word and document-topic distribution, while \citet{Hu2020NeuralTM} uses CycleGAN \cite{Zhu2017UnpairedIT} for unsupervised transfer between document-word and document-topic distribution.

Hierarchical topic models \cite{viegas-etal-2020-cluhtm} utilize relationships among the latent topics. Supervised topic models have been explored previously where the topic model is trained through human feedback \cite{Kumar2019WhyDY} or with a task specific network simultaneously such that topic extraction is guided through task labels \cite{Pergola2019TDAMAT, Wang2020NeuralTM}. \citet{card-etal-2018-neural} leverages document metadata but without metadata their method is same as ProdLDA which is our baseline. Topic modeling on document networks has been done leveraging relational links between documents \cite{Zhang2020TopicMO, zhou-etal-2020-neural}. However our problem setting is completely different, we extract topics from documents in unsupervised way where document links/metadata/labels either don’t exist or are not used to extract the topics. 

Some very recent works use pre-trained BERT \cite{devlin-etal-2019-bert} either to leverage improved text representations \cite{Bianchi2020PretrainingIA, sia-etal-2020-tired} or to augment topic model through knowledge distillation \cite{Hoyle2020ImprovingNT}. \citet{Zhu2020ANG} and \citet{dieng-etal-2020-topic} jointly train words and topics in a shared embedding space. However, we train topic-word distribution as part of our model, embed it using word embeddings being learned and use resultant topic embeddings to perform attention over sequentially processed tokens. iDocNade \cite{gupta2019document} is an autoregressive topic model for short texts utilizing pre-trained embeddings as distributional prior. However, it attains poorer topic coherence than ProdLDA and GNB-NTM as shown in \citet{wu-etal-2020-neural}.

Some works have attempted to use other prior distributions such as \citet{Zhang2018WHAIWH} uses the Weibull prior, \citet{pmlr-v2-thibaux07a} uses the beta distribution. Gamma Negative Binomial-Neural Topic Model (GNB-NTM) \cite{wu-etal-2020-neural} is one of the recent neural variational topic models which attempt to combine VI with mixed counting models. Mixed counting models can better model hierarchically dependent and over-dispersed random variables while implicitly introducing non-negative constraints in topic modeling. GNB-NTM uses reparameterization of Gamma distribution and Gaussian approximation of Poisson distribution. We use their model as a baseline for our work.

Topic models have been used with sequence encoders such as LSTM in applications like user activity modeling \cite{Zaheer2017LatentLA}. \citet{dieng2016topicrnn} employs an RNN to detect stop words and merges its output with document-topic vector for next word prediction. \citet{Gururangan2019VariationalPF} uses a VAE pre-trained through topic modeling to perform text classification. We perform document classification and compare our model's accuracy with the accuracy of VAE based and other topic models. LTMF \cite{jin-etal-2018-combining} combines text features processed through an LSTM with a topic model for review based recommendations. Fundamentally different from these, we use topic-word distribution to attend on sequentially processed tokens via novel topic guided attention for performing variational inference, learning better document-topic features and improving topic modeling.

A key application of topic models is supervised keyphrase generation. Some of the existing neural keyphrase generation methods include SEQ-TAG \cite{zhang-etal-2016-keyphrase} based on sequence tagging, SEQ2SEQ-CORR \cite{chen-etal-2018-keyphrase} based on seq2seq model without copy mechanism and SEQ2SEQ-COPY \cite{meng-etal-2017-deep} which additionally uses copy mechanism. Topic-Aware Keyphrase Generation (TAKG) \cite{wang-etal-2019-topic} is a seq2seq based neural keyphrase generation framework for social media language. TAKG uses a neural topic model in \citet{miao17a} and a keyphrase generation (KG) module which is conditioned on latent document-topic vector from the topic model. We adapt our proposed topic model to TAKG to improve keyphrase generation and discuss it in detail later in the Experiments section.

\begin{figure*}[t]
\centering
\includegraphics[width=\textwidth,height=3in]{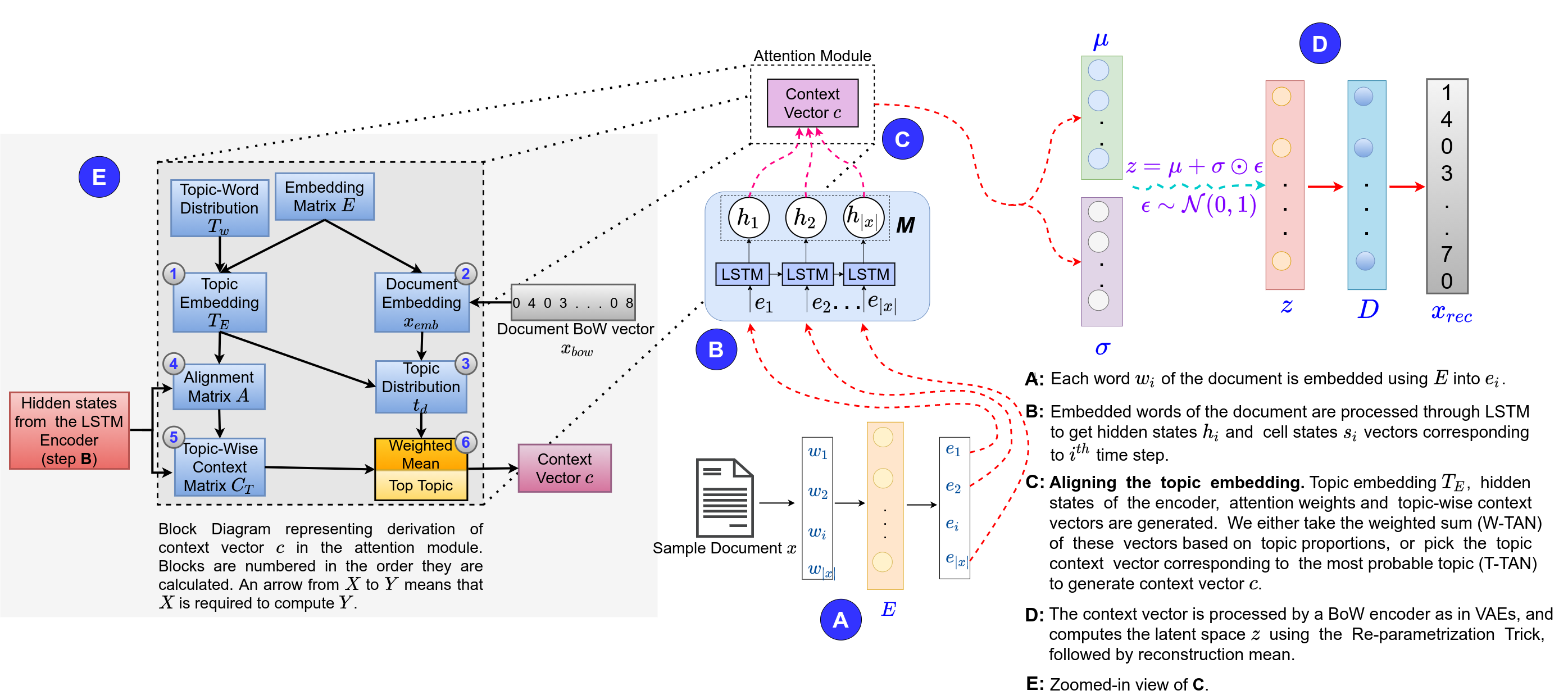} 
\caption{\textbf{A-E}: Architecture of TAN-NTM showing flow of document processing through it. Document, being embedded using embedding layer, is processed by LSTM, yielding hidden states on which TAN attends in a topic aware manner. The resultant context vector is used to perform variational inference and processed through a BoW decoder as in VAEs. Attention Module E (zoomed in view of C) computes the blocks in the mentioned order \textbf{1-6}.}
\label{fig:TAN-arch}
\end{figure*} 


\section{Background}

LDA is a generative statistical model and assumes that each document is a distribution over a fixed number of topics (say $K$) and that each topic is a distribution of words over the entire vocabulary. LDA proposes an iterative process of document generation where for each document $d$, we draw a topic distribution $\theta$ from $Dirichlet(\alpha)$ distribution. For each word in $d$ at index $i$, we sample a topic $\mathbf{t}_i$ from $Multinomial(\theta)$ distribution. $\mathbf{w}_i$ is sampled from $p(\mathbf{w}_i|\mathbf{t}_i,\beta)$ distribution which is a multinomial probability conditioned on topic $\mathbf{t}_i$. Given the document corpus and the parameters $\alpha$ and $\beta$, we need the joint probability distribution of a topic mixture \textbf{$\theta$}, a set of $K$ topics \textbf{t}, and a set of $n$ words \textbf{w}. This is given analytically by an intractable integral. The solution is to use Variational Inference wherein this problem is converted into an optimization problem for finding various parameters that minimize the KL divergence between the prior and the posterior distribution.

This idea is leveraged at scale by the use of Variational Autoencoders. The encoder processes BoW vector of the document $\mathbf{x}_{bow}$ by an MLP (Multi Layer Perceptron) which then forks into two independently trainable layers to yield $\mathbf{z}_{\mu}$ \& $\mathbf{z}_{\log \sigma^2}$. Then a re-parametrization trick is employed to sample the latent vector $\mathbf{z}$ from a logistic-normal distribution (resulting from an approximation of Dirichlet distribution). This is essential since back-propagation through a sampling node is infeasible. $\mathbf{z}$ is then used by decoder's single dense layer \textbf{D} to yield the reconstructed BoW $\mathbf{x}_{rec}$. The objective function has two terms: \textbf{(a) Kullback–Leibler (KL) Divergence Term} - to match the variational posterior over latent variables with the prior and \textbf{(b) Reconstruction Term} - categorical cross entropy loss between $\mathbf{x}_{bow}$ \& $\mathbf{x}_{rec}$.
$$L_{NTM} = D_{KL}(p(\mathbf{z})\  ||\  q(\mathbf{z}|\mathbf{x}))-\mathbb{E}_{q(\mathbf{z}|\mathbf{x})}[p(\mathbf{x}|\mathbf{z})]$$
Our methodology improves upon the document encoder and introduces a topic guided attention whose output is used to sample $\mathbf{z}$. We use the same formulation of decoder as used in ProdLDA.


\section{Methodology}
In this section, we describe the details of our framework where we leverage the topic-word distribution to perform topic guided attention over tokens in a document. Given a collection $\mathcal{C}$ with $|C|$ documents $\mathbf{\{ x_1, x_2,.., x_{|C|} \}}$, we process each document $\mathbf{x}$ into BoW vector $\mathbf{x}_{bow} \in \mathbb{R}^{|V|}$ and as a token sequence $\mathbf{x}_{seq}$, where V represents the vocabulary. As shown in step A in figure \ref{fig:TAN-arch}, each word $w_j \in \mathbf{x}_{seq}$ is embedded as $\mathbf{e}_j \in \mathbb{R}^{E}$ through an embedding layer $\mathbf{E} \in \mathbb{R}^{|V|\times E}$ ($E =$ Embedding Dimension) initialised with GloVe \cite{pennington2014glove}. The embedded sequence $\{{e}_{j}\}_{j=1}^{\mathbf{|x|}}$, where $\mathbf{|x|}$ is the number of tokens in $\mathbf{x}$, is processed through a sequence encoder LSTM \cite{10.1162/neco.1997.9.8.1735} to obtain the corresponding hidden states $\mathbf{h}_j \in \mathbb{R}^H$ and cell states $\mathbf{s}_j \in \mathbb{R}^H$ (step B in figure \ref{fig:TAN-arch}):
$$\mathbf{h}_j,\mathbf{s}_j=f_{LSTM} (\mathbf{e}_j,(\mathbf{h}_{j-1},\mathbf{s}_{j-1}))$$


 where $H$ is LSTM's hidden size. We construct a memory bank $\mathbf{M} = \langle \mathbf{h}_1,\mathbf{h}_2,...,\mathbf{h_{|x|}} \rangle$ which is then used to perform topic-guided attention (step C in figure \ref{fig:TAN-arch}). The output vector of the attention module is used to derive prior distribution parameters $\mathbf{z}_{\mu}$ \& $\mathbf{z}_{\log \sigma^2}$ (as in VAE) through two linear layers. Using the re-parameterisation trick, we sample the latent document-topic vector $\mathbf{z}$, which is then given as input to BoW decoder linear layer $\mathbf{D}$ that outputs the reconstructed BoW $\mathbf{x}_{rec}$ (step D in figure \ref{fig:TAN-arch}). Objective function is same as in VAE setting, involving a reconstruction loss term between $\mathbf{x}_{rec}$ \& $\mathbf{x}_{bow}$ and KL divergence between the prior (laplace approximation to Dirichlet prior as in ProdLDA) and posterior. We now discuss the details of our Topic Attention Network. 

\subsection{TAN: Topic Attention Network}
We intend the model to attend on document words in a manner such that the resultant attention is distributed according to the semantics of the topics relevant to the document. We hypothesize that this can enable the model to encode better document features while capturing the underlying latent document-topic representations. The topic-word distribution $\mathbf{T_w}$ represents the affinity of each topic towards words in the vocabulary (which is used to interpret the semantics of each topic). Therefore, we factor $\mathbf{T_w} \in \mathbb{R}^{K \times |V|}$ into the attention mechanism, where K denotes the number of topics. The topic-aware attention encoder and topic-word distribution influence each other during training which consequently results in convergence to better topics as discussed in detail in Experiments section.

Specifically, we perform attention on document sequence of tokens for each topic using the embedded representation of the topics $\mathbf{T_E} \in \mathbb{R}^{K\times E}$:

$$\mathbf{T_E}= \mathbf{T_w}\mathbf{E} , \ \ \ \ \textrm{[topic embeddings]}$$
$$\mathbf{T_w}=\mathrm{softmax}(\mathbf{D}), \ \ \ \  \textrm{[topic-word distribution]}$$


 where $\mathbf{D} \in \mathbb{R}^{K \times V}$ is the decoder layer which is used to reconstruct $\mathbf{x}_{bow}$ from the sampled latent document-topic representation $\mathbf{z}$ as the final step D in Figure \ref{fig:TAN-arch}. The topic embeddings are then used to determine the attention alignment matrix $\mathbf{A} \in \mathbb{R}^{\mathbf{|x|}\times K}$ between each topic $k \in \{1, 2, ..., K\}$ and words in the document such that:
$$\mathbf{A}_{jk}=\frac{ \mathrm{exp} (score ((\mathbf{T_E})_{k}, \mathbf{h}_j) ) }{ \sum_{j'=1}^{\mathbf{|x|}} \mathrm{exp} (score ((\mathbf{T_E})_{k}, \mathbf{h}_{j'}) )},$$
$$score((\mathbf{T_E})_k, \mathbf{h}_j)=\mathbf{v_A}^\top  \mathrm{tanh}( \mathbf{W_A}[(\mathbf{T_E})_k \mathbf{;} \mathbf{h}_j])$$
\indent where $\mathbf{v_A} \in \mathbb{R}^{P}$, $\mathbf{W_A} \in \mathbb{R}^{P \times (E+H)}$, $(\mathbf{T_E})_k \in \mathbb{R}^E$  is the embedded representation of the $k^{th}$ topic and $;$ is the concatenation operation. We then determine topic-wise context vector corresponding to each topic as: 
$$\mathbf{C_T}= \sum_{j=1}^{\mathbf{|x|}} \mathbf{A}_j \otimes \mathbf{h}_j,\ \ \ \ \textrm{[topic-wise context matrix]}$$ 



 where $\otimes$ denotes outer product. Note that $\mathbf{A}_j$ $\in \mathbb{R}^{K}$ ($j^{th}$ row of matrix $\mathbf{A}$) is a $K$ - dimensional vector and $\mathbf{h}_j$ is a $H$ - dimensional vector, therefore $\mathbf{A}_j \otimes \mathbf{h}_j$ for each $j$ yields a matrix of order $K \times H$, hence $\mathbf{C_T} \in \mathbb{R}^{K\times H}$. The final aggregated context vector $\mathbf{c}$ is computed as a weighted average over all rows of $\mathbf{C_T}$ (each row representing each topic specific context vector) with document-topic proportion vector $\mathbf{t_d}$ as weights: 
$$\mathbf{c}= \sum_{k=1}^{K} (\mathbf{t_d})_i(\mathbf{C_T})_k$$
\indent where, $(\mathbf{t_d})_k$ is a scalar, $(\mathbf{C_T})_k \in \mathbb{R}^{H}$ denotes the $k^{th}$ row of matrix $\mathbf{C_T}$ \& $\mathbf{t_d}$ is the document-topic distribution which signifies the topic proportions in a document. To compute it, we first normalize the document BoW vector $\mathbf{x}_{bow}$ and embed it using the embedding matrix $\mathbf{E}$, followed by multiplication with topic embedding $\mathbf{T_E} \in \mathbb{R}^{K \times E}$:
$$\mathbf{x}_{norm}=\frac{\mathbf{x}_{bow}}{\sum_{i=1}^{|V|}(\mathbf{x}_{bow})_i }, \ \  \textrm{[normalized BoW]}$$
$$\mathbf{x}_{emb}=\mathbf{x}_{norm}^\top E, \ \ \ \ \ \ \ \ \ \  \textrm{[document embedding]}$$
$$\mathbf{t_d}=\mathrm{softmax}(\mathbf{T_E} \ \mathbf{x}_{emb}), \textrm{[document-topic dist.]}$$
\indent where $\mathbf{x_{norm}} \in \mathbb{R}^{|V|}$, $\mathbf{x_{emb}} \in \mathbb{R}^{E}$ \& $\mathbf{t_d} \in \mathbb{R}^{K}$. The context vector $\mathbf{c}$ is the output of our topic guided attention module which is then used for sampling the latent documents-topic vector followed by the BoW decoding as done in traditional VAE based topic models.\par 
We call this framework as Weighted-TAN or W-TAN where the context vector $\mathbf{c}$ is a weighted sum of topic-wise context vectors. We also propose another model called Top-TAN or T-TAN where we use context vector of the topic with largest proportion in $\mathbf{t_d}$ as $\mathbf{c}$. It has been experimentally observed that doing so yields a model which generates more coherent topics. First, we find the index $m$ of most probable topic in $\mathbf{t_d}$. The context vector $\mathbf{c}$ is then the row corresponding to index $m$ in matrix $\mathbf{C_T}$.

\section{Experiments}

\subsection{Datasets}
\textbf{1. Topic Quality:} We evaluate and compare quality of our proposed topic model on three benchmark datasets - 20Newsgroups (20NG)\footnote{ \href{https://github.com/akashgit/autoencoding_vi_for_topic_models/tree/master/data/20news_clean}{Data link} for 20NG dataset} \cite{lang1995newsweeder}, AGNews \cite{zhang2015character} and Yelp Review Polarity (YRP)\footnote{ \href{https://drive.google.com/drive/u/0/folders/0Bz8a_Dbh9Qhbfll6bVpmNUtUcFdjYmF2SEpmZUZUcVNiMUw1TWN6RDV3a0JHT3kxLVhVR2M}{Data link} for AGNews and YRP datasets} - which are of varying complexity and scale in terms of number of documents, vocabulary size and average length of text after preprocessing\footnote{We provide our detailed preprocessing steps in Appendix \ref{appendix:preprocessing} and release \href{https://drive.google.com/file/d/1pNrAH-PAwWQdpoE0eVPLG1sLRO7buvw_/view?usp=sharing}{processed data} to standardise it.}. Table \ref{table:data} summarises statistics related to these datasets used for evaluating topics quality.


\begin{table}[hbt!]
 \centering
 \resizebox{220pt}{!}{%
 \begin{tabular}{ccccc}

 \hline 
  Dataset & \# Train & \# Test & vocab & avg.doc.len. \\
  \hline 
  20NG & 11259 & 7488 & 1995 & 88.06 \\
  
  AGNews & 96000 & 7600 & 27881 & 22.72 \\
  
  YRP & 447873 & 38000 & 20001 & 54.46 \\
  \hline
\end{tabular}}
\caption{Datasets used for evaluating topic quality}
\label{table:data}
\end{table}


\noindent\textbf{2. Keyphrase Generation:} Neural Topic Model (NTM) has been used to improve the task of supervised keyphrase generation \cite{wang-etal-2019-topic}. To further highlight the efficacy of our proposed encoding framework in providing better document-topic vectors, we modify encoder module of NTM with our proposed TAN-NTM and compare the performance on \textbf{StackExchange} and \textbf{Weibo} Datasets\footnote{The dataset details can be found in the \href{https://www.aclweb.org/anthology/P19-1240.pdf}{baseline paper}}. 




\subsection{Implementation and Training Details}\label{section:impl-and-train}


Documents in AGNews are padded upto a maximum length of $50$, while those in 20NG and YRP are padded upto $200$ tokens. Documents with longer lengths are truncated. These values were chosen such that $\sim80-99\%$ of all documents in each dataset were included without truncation. We use batch size of 100, Adam Optimizer \cite{Kingma2015AdamAM} with $\beta_1=0.99$, $\beta_2=0.999$ and $\epsilon = 10^{-8}$ and train each model for $200$ epochs. For all models except T-TAN, learning rate was fixed at $\textbf{0.002}\ ([0.001,\ 0.003],\ 5)$\footnote{$\textbf{V}\ ([a,\ b],\ t)$ means $t$ values from $[a, b]$ range tried for this hyper-parameter, of which $\textbf{V}$ yielded best NPMI coherence.}. T-TAN converges relatively faster than other models, therefore for smooth training, we decay its learning rate every epoch using exponential staircase scheduler with initial learning rate = 0.002 and decay rate = 0.96. The number of topics $K$ = $50$, a value widely used in literature. We perform hyper-parameter tuning manually to determine the hidden dimension value of various layers: $E$ = $\textbf{200}\ ([100,\ 300],\ 5)$, $H$ = $\textbf{450}\ ([300,\ 900],\ 10)$ and $P$ = $\textbf{350}\ ([10,\ 400],\ 10)$. The weight matrices of all dense layers are Xavier initialized, while bias terms are initialized with zeros. All our proposed models and baselines are trained on a machine with 32 virtual CPUs, single NVIDIA Tesla $V100$ GPU and 240 GB RAM.

\subsection{Comparison with baselines}
We compare our TAN-NTM with various baselines in table \ref{table:coh_res} that can be enumerated as (please refer to introduction and related work for their details): \\
\textbf{1)} \textbf{LDA (C.G.)}: Statistical method  \cite{mccallum2002mallet} which performs LDA using collapsed Gibbs\footnote{\url{https://pypi.org/project/lda/}} sampling. \\
\textbf{2)} \textbf{ProdLDA} and \textbf{3)} \textbf{NVLDA} \cite{srivastava2017autoencoding}: Neural Variational Inference methods which use approximation to Dirichlet prior\footnote{\href{https://github.com/akashgit/autoencoding_vi_for_topic_models}{Code} for ProdLDA and NVLDA}. \\
\textbf{4)} \textbf{W-LDA} \cite{nan-etal-2019-topic} which is a non variational inference based neural model using wassestein autoencoder\footnote{\url{https://github.com/awslabs/w-lda}}. \\
\textbf{5)} \textbf{NB-NTM} and \textbf{6)} \textbf{GNB-NTM}: Methods using negative binomial and gamma negative binomial distribution as priors for topic discovery\footnote{We thank authors for providing \href{https://github.com/mxiny/NB-NTM}{code} and parameter info.}\cite{wu-etal-2020-neural} respectively.

\begin{table}[h]
 \centering
 \begin{tabular}{cccc}

 \hline 
  Method  & 20NG & AGNews & YRP  \\
  \hline 
  LDA(C.G) & 0.139 & 0.202 & 0.114  \\
  NVLDA & 0.2 & 0.216 & 0.165  \\
  ProdLDA & 0.268 & 0.322 & 0.165 \\
  W-LDA & 0.227 & 0.262 & 0.25  \\
  NB-NTM & 0.165 & 0.31 & 0.224 \\
  GNB-NTM & 0.206 & 0.312 & 0.241  \\
  \hline
  \hline
  \textbf{W-TAN} (ours) & 0.261 & 0.327 & 0.232  \\
  \textbf{T-TAN} (ours) & \textbf{0.296} & \textbf{0.369} & \textbf{0.272} \\
  \hline
\end{tabular}
\caption{NPMI coherence (determined using top 10 words of each topic) comparison on 50 topics between baselines and our proposed W-TAN and T-TAN on different datasets. It can be seen that T-TAN achieves significantly better scores on all the datasets.}
\label{table:coh_res}
\end{table}

 We could not compare with other methods whose official error-free source code is not publicly available yet. We train and evaluate the baseline methods on same data as used for our method using NPMI coherence\footnote{\href{https://github.com/jhlau/topic_interpretability}{Repo} used to calculate NPMI. Please refer to Appendix \ref{appendix:eval-metric} for a detailed discussion on choice of evaluation metric.} \cite{aletras2013evaluating}. It computes the semantic relatedness between top $L$ words in a given topic through determining similarity between their word embeddings trained over the corpus used for topic modeling and reports average over topics. For W-LDA, we refer to their original paper to select dataset specific hyper-parameter values while training the model. As can be seen in table \ref{table:coh_res}, our proposed T-TAN model performs significantly better than previous topic models uniformly on all datasets achieving a better NPMI (measured on a scale of -1 to 1) by a margin of 0.028 (10.44\%) on 20NG, 0.047 (14.59\%) on AGNews and 0.022 (8.8\%) on YRP, where percentage improvements are determined over the best baseline score. Even though W-TAN does not uniformly performs better than all baselines on all datasets, it achieves better score than all baselines on AGNews and performs comparably on remaining two datasets.


For a more exhaustive comparison, we also evaluate our model’s performance on 20NG dataset (which is the common dataset with GNB-NTM \cite{wu-etal-2020-neural}) using the NPMI metric from GNB-NTM’s code. The NPMI coherence of our model using their criteria is 0.395 which is better than GNB-NTM’s score of 0.375 (as reported in their paper). However, we would like to highlight that GNB-NTM’s computation of NPMI metric uses relaxed window size, whereas the metric used by us \cite{lau-etal-2014-machine} uses much stricter window size while determining word co-occurrence counts within a document. \citet{lau-etal-2014-machine} is a much more common and widely used way of computing the NPMI coherence and evaluating topic models.

\subsubsection{Document Classification}
In addition to evaluating our framework in terms of topic coherence, we also compare it with the baselines on the downstream task of document classification. Topic models have been used as text feature extractors to perform classification \cite{nan-etal-2019-topic}. We analyse the quality of encoded document representations and predictive capacity of latent document-topic features generated by our model and compare it with existing topic models\footnote{Our aim is to analyse document-topic features among topic models only and not to compare with other non-topic model based generic text classifiers.}. We train the topic model setting number of topics to 50 and freeze its weights. The trained topic model is then used to infer latent document-topic features. We then separately train a single layer linear classifier through cross entropy loss on the training split using the document-topic vectors as input and Adam optimizer at a learning rate of 0.01. 

\begin{table}[h]
 \centering
 
 \begin{tabular}{cccc}

 \hline 
    Method  & 20NG & AGNews  & YRP \\
  \hline
    LDA(C.G.) & 51.29 & 84.78  & 86.85 \\
  ProdLDA & 21.33 & 82.65  & 77.73 \\
  NTM-R & 43.34 & 85.67  & 86.16 \\
  W-LDA & 43.08 & 85.29  & 85.63 \\
  NB-NTM & 57.38 & 86.67  & \textbf{87.51} \\
  GNB-NTM & 57.16 & 85.34  & 84.55 \\
  \hline
  \hline
  \textbf{T-TAN} (ours) & \textbf{60.44} & \textbf{88.1} & 87.38 \\
  \hline
  \textbf{T-TAN} (ours) & \textbf{64.36} & \textbf{89.78}  & \textbf{88.9} \\
  (context vector) & & & \\
  \hline
  
\end{tabular}
\caption{Comparison of accuracy between different topic models on document classification. We perform two experiments with T-TAN: using document-topic vector ($2^{nd}$ to last row) and context vector (last row).}
\label{table:class}
\end{table}

We report classification accuracy on the test split of 20NG, AGNews and YRP datasets (comprising of 20, 4 and 2 classes respectively) in Table \ref{table:class}. The document-topic features provided by T-TAN achieve best accuracy on AGNews (1.43\% improvement over most performant baseline) with most significant improvement of 3.06\% on 20NG which shows our model learns better document features. T-TAN performs almost the same as the best baseline on YRP. Further, to analyse the predictive performance of top topic attention based context vector, we use it instead of latent document-topic vector to perform classification which further boosts accuracy leading to an improvement of $\sim$6.9\% on 20NG, $\sim$3.1\% on AGNews and $\sim$1.3\% on YRP datasets over the baselines.





\subsubsection{Running Time Analysis}
We compare the running time of our method with baselines in terms of average time taken (in seconds) for performing a forward pass through the model, where the average is taken over 10000 passes. Our TAN-NTM (implemented in tensorflow) takes 0.087s, 0.027s and 0.093s on 20NG, AGNews and YRP datasets respectively. Since TAN-NTM processes the input documents as a sequence of tokens through an LSTM, its running time is proportional to the document lengths which vary according to the dataset. The running time for baseline methods are: ProdLDA - 0.012s (implemented in tensorflow), W-LDA - 0.003s (implemented in mxnet) and GNB-NTM - 0.003s (implemented in pytorch). For baseline methods, we have used their original code implementations. We found that the running time of baseline models is independent of the dataset. This is because they use the Bag-of-Words (BoW) representation of the documents. The sequential processing in TAN-NTM is the reason for increased running time of our models compared to the baselines. In the case of AGNews, since the documents are of lesser lengths than 20NG and YRP, the running time of our TAN-NTM is relatively less for AGNews. Further, the running time of other ablation variants (introduced in section \ref{section:ablations}) of our method on 20NG, AGNews and YRP datasets respectively are: 1) only LSTM - 0.083s, 0.033s and 0.091s ; 2) vanilla attn - 0.088s, 0.037s and 0.095s.

\subsection{Ablation Studies}\label{section:ablations}
In this section, we compare the performance of different variants of our model namely, \textbf{1)} \textbf{only LSTM}: final hidden state is used to derive sampling parameters $\mathbf{z}_{\mu}$ \& $\mathbf{z}_{\log \sigma^2}$, \textbf{2)} \textbf{vanilla attn}: final hidden state (w/o topic-word distribution) is used as query to perform attention  \cite{bahdanau2014neural} on LSTM outputs such that context vector $\mathbf{z}$ is used for VI, \textbf{3)} \textbf{W-TAN}: Weighted Topic Attention Network, \textbf{4)} \textbf{T-TAN}: Top Topic Attention Network and \textbf{5)} \textbf{T-TAN w/o (without) GloVe}: embedding layer in T-TAN is randomly initialised. 

Table \ref{table:abl} compares the topic coherence scores of these different ablation methods on 20NG, AGNews and YRP. As can be seen, applying attention performs better than simple LSTM model. The weighted TAN performs better than vanilla attention model, however, T-TAN uniformly provides the best coherence scores across all the datasets compared to all other methods. This shows that performing attention corresponding to the most prominent topic in a document results in more coherent topics. Further, we perform an ablation to study the effect of using pre-trained embeddings for T-TAN where it can be seen using Glove for initialising word embeddings results in improved NPMI as compared to training T-TAN initialised with random uniform embeddings (T-TAN w/o GloVe)\footnote{We also trained embeddings from scratch for other variants but coherence score remained unaffected.}.

\begin{table}[h]
 \centering
 \begin{tabular}{cccc}

 \hline 
  Method  & 20NG & AGNews & YRP  \\
  \hline 
  only LSTM & 0.247 & 0.202 & 0.092  \\
  vanilla attn & 0.289 & 0.244 & 0.18  \\
  W-TAN & 0.261 & 0.327 & 0.232 \\
  T-TAN & \textbf{0.296} & \textbf{0.369} & \textbf{0.272} \\
  T-TAN w/o GloVe & 0.274 & 0.344 & 0.248  \\
  \hline
\end{tabular}
\caption{Comparison of NPMI coherence between ablation variants of our method for K=50 topics.}
\label{table:abl}
\end{table}

\begin{table*}[t]
 \centering
 \begin{tabular}{c|ccc|ccc}

 \hline 
     & \multicolumn{3}{c|}{StackExchange}  & \multicolumn{3}{c}{Weibo}  \\
  Method & F1@3 & F1@5 & MAP & F1@1 & F1@3 & MAP \\
  \hline
  TAKG (baseline)  & 32.931 & 28.731 & 34.925 & 34.584 & 24.309 & 40.994\\
  \hline
  \hline
  TAKG with W-TAN (ours)  & \textbf{33.521} & \textbf{29.802} & \textbf{35.929} & \textbf{35.616} & \textbf{25.651} & \textbf{42.68}\\
  TAKG with T-TAN (ours) & 33.15 & 29.118 & 35.26 & 34.813 & 24.65 & 41.261\\
  \hline
 \end{tabular}
\caption{F1@k and MAP (Mean average precision) comparison between baseline (TAKG) and our proposed topic model based encoder for topic guided supervised keyphrase generation. The metrics measure overlap between ground truth and top K generated keyphrases factoring in rank of keyphrases generated through beam search.}
\label{table:takg}
\end{table*}





\subsection{Qualitative Analysis}
\begin{figure}[h]
\centering
\includegraphics[scale=0.30]{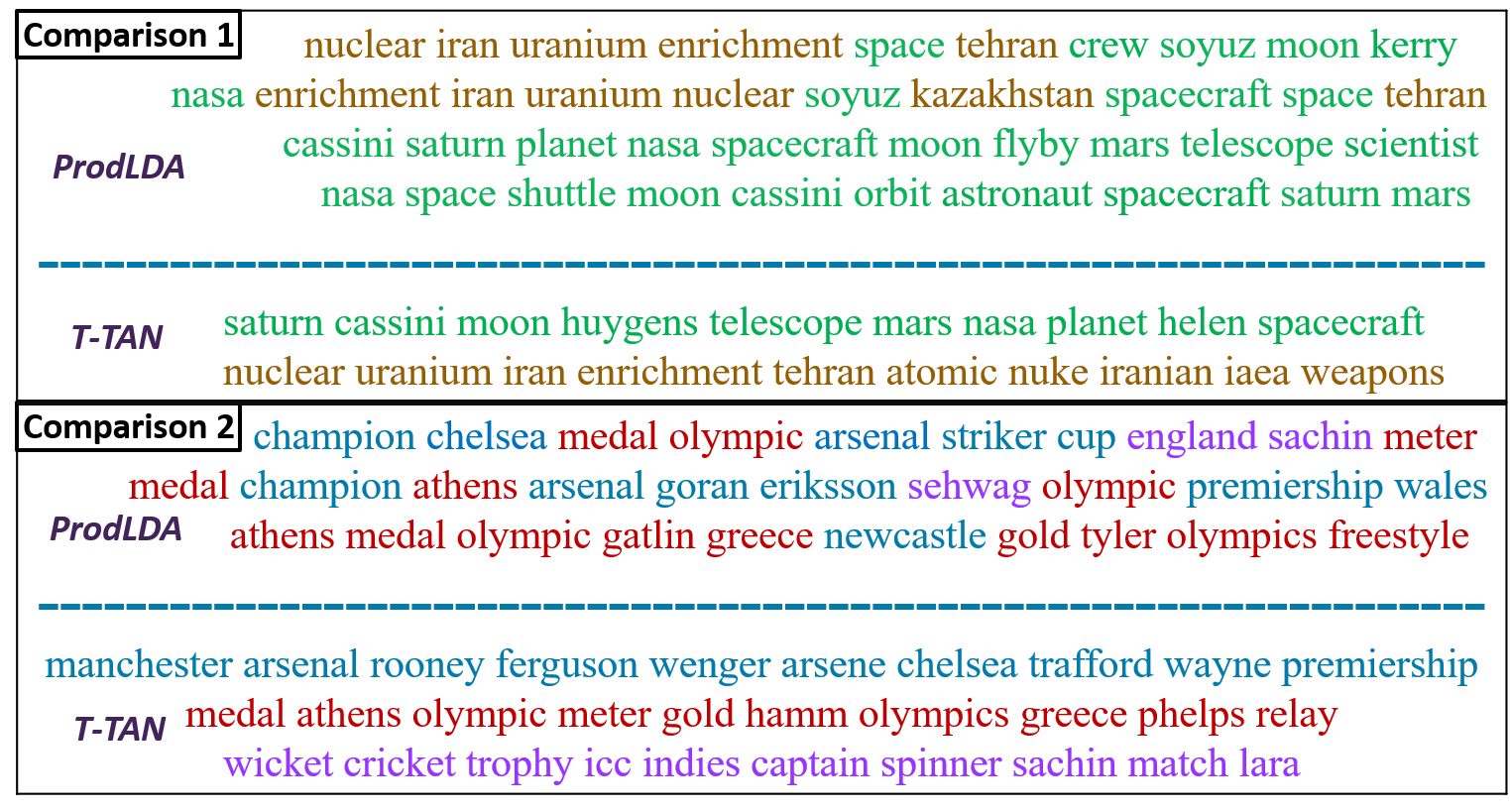}

\caption{Two comparisons of corresponding topics (one topic per line) from ProdLDA and T-TAN. Words having similar meaning are highlighted in same colour. The topics of ProdLDA are inter-mixed and incoherent while those of T-TAN are unmixed and coherent.}
\label{fig:qual_anal_topics}
\end{figure}

To verify performance of T-TAN qualitatively, we display few topics generated by ProdLDA and T-TAN on AGNews in Figure \ref{fig:qual_anal_topics}. ProdLDA achieves best score among baselines on AGNews. Consider comparison 1 in Figure \ref{fig:qual_anal_topics}: ProdLDA produces four topics corresponding to space, mixing them with nuclear weapons, while T-TAN produces two separate topics for both of these concepts. In second comparison, we see that ProdLDA has problems distinguishing between closely related topics (football, olympics, cricket) and mixes them while T-TAN produces three coherent topics.

\subsection{TAKG: Topic Aware Keyphrase Generation}
We further analyse the impact of our proposed framework on another downstream task where the task specific model is assisted by the topic model and both can be trained in an end-to-end manner. For this, we discuss TAKG \cite{wang-etal-2019-topic} and how our proposed topic model encoder can be adapted to achieve better performance on supervised keyphrase generation from textual posts. TAKG\footnote{We use their code and data (\href{https://github.com/yuewang-cuhk/TAKG}{link}) to conduct experiments.} comprises of two sub-modules: \textbf{(1)} a topic model based on NVDM-GSM (as discussed in Introduction) using BoW as input to the encoder and \textbf{(2)} a Seq2Seq based model for keyphrase generation. Both modules have an encoder and a decoder of their own. Keyphrase generation module uses sequence input which is processed by bidirectional GRU \cite{cho-etal-2014-learning} to encode input sequence. The keyphrase generation decoder uses unidirectional GRU which attends on encoder outputs and takes the latent document-topic vector from the topic model as input in a differentiable manner. Since topic model trains slower than keyphrase generation module, the topic model is warmed up for some epochs separately and then jointly trained with keyphrase generation. Please refer to original paper \cite{wang-etal-2019-topic} for more details.



We adapted our proposed topic model framework by changing the architecture of encoder in the topic model of TAKG, replacing it with W-TAN and T-TAN. The change subsequently results in better latent document-topic representation depicted by better performance on keyphrase generation as shown in Table \ref{table:takg} where the improved topic model encoding framework results in $\sim$1-2\% improvement in F1 and MAP (mean average precision) on StackExchange and Weibo datasets compared to TAKG. Here, even though TAKG with T-TAN  performs marginally better than the baseline, TAKG with W-TAN uniformly performs much better.
 


\section{Conclusion}

In this work, we propose Topic Attention Network based Neural Topic Modeling framework: TAN-NTM to discover topics in a document corpus by performing attention on sequentially processed tokens in a topic guided manner. Attention is performed effectively by factoring Topic-word distribution (TWD) into attention mechanism. We compare different variants of our method through ablations and conclude that processing tokens sequentially without attention or applying attention without TWD gives inferior performance. Our TAN-NTM model generates more coherent topics compared to state-of-the-art topic models on several benchmark datasets. Our model encodes better latent document-topic features as validated through better performance on document classification and supervised keyphrase generation tasks. As future work, we would like to explore our framework with other sequence encoders such as Transformers, BERT etc. for topic modeling.






\bibliographystyle{acl_natbib}
\bibliography{acl2021}

\clearpage
\appendix
\begin{Large}
\begin{flushleft}
\textbf{Appendices}
\end{flushleft}
\end{Large}
\section{Further Implementation Details}
\subsection{Preprocessing}\label{appendix:preprocessing}
For 20NG dataset, we used its preprocessed version downloaded from ProdLDA's \cite{srivastava2017autoencoding} repository\footnote{ \href{https://github.com/akashgit/autoencoding_vi_for_topic_models/tree/master/data/20news_clean}{Data link} for 20NG dataset}, whereas AGNews and YRP datasets were downloaded from this\footnote{ \href{https://drive.google.com/drive/u/0/folders/0Bz8a_Dbh9Qhbfll6bVpmNUtUcFdjYmF2SEpmZUZUcVNiMUw1TWN6RDV3a0JHT3kxLVhVR2M}{Data link} for AGNews and YRP datasets} link. These two datasets contain \textbf{train.csv} and \textbf{test.csv} files. The csv files of YRP contain a document body only, whereas the csv files for AGNews contain a document title as well as a document body. For uniformity, we concatenate the title and body in the csv files of AGNews and keep it as a single field. The documents from \textbf{train.csv} and \textbf{test.csv} are then read into \texttt{train} and \texttt{test} lists which are passed to \textsc{preprocess} function of Algorithm \ref{algorithm:preprocess} for preprocessing.\\
\\
Stepwise working of Algorithm \ref{algorithm:preprocess} is expained in the following points:
\begin{itemize}

\item Before invoking the \textsc{preprocess} function, we initialize the data sampler by a fixed seed so that preprocessing yields the same result when run multiple times. 

\item For each dataset, we randomly sample \texttt{tr\_size} documents (as mentioned in Table \ref{table:preprocess-params}) from the \texttt{train} list in step 2. These values of \texttt{tr\_size} are taken from Table 1 of W-LDA paper \cite{nan-etal-2019-topic}. Note that \textbf{\# Train} in Table \ref{table:data} represents the number of training documents after preprocessing. Of the \texttt{tr\_size} documents, some documents may be removed during preprocessing, therefore \textbf{\# Train} may be less than \texttt{tr\_size}.

\item In steps 3 through 8, we prune the \texttt{train} and \texttt{test} documents by invoking the \textsc{prune\_doc} function from Algorithm \ref{algorithm:encode-prune}. First, we remove the control characters from the documents viz. \lq \textbackslash n\rq, \lq \textbackslash t\rq, and \lq \textbackslash r\rq \ (For YRP, we additionally remove \textbf{\lq \textbackslash \textbackslash t\rq}, \textbf{\lq \textbackslash \textbackslash n\rq}, and \textbf{\lq \textbackslash \textbackslash r\rq}). Next, we remove the numeric tokens\footnote{Fully numeric tokens e.g. \lq1487\rq, \lq1947\rq, etc. are removed, whereas partially numeric tokens e.g. \lq G47\rq, \lq DE1080\rq, etc. are retained.} from the documents, convert them to lowercase and lemmatize each of their tokens using the NLTK's \cite{BirdKleinLoper09} WordNetLemmatizer. Finally, we remove punctuations\footnote{Any of the following 32 characters is regarded as a punctuation !"\#\$\%\&'()*+,-./:;\textless=\textgreater?@[\textbackslash]\^{}\_\textasciigrave\{\textbar\}$\sim$} and tokens containing any non-ASCII character. 

\item In steps 9 through 15, we construct the vocabulary \texttt{vocab}, which is a mapping of each token to its occurrence count among the pruned training documents \texttt{tr\_pruned}. We only count a token if it is not an English stopword\footnote{Gensim's \cite{rehurek_lrec} list of English stopwords is used.} and its length is between 3 and 15 (inclusive).

\item Steps 16 through 19 filter the \texttt{vocab} by removing tokens whose total occurrence count is less than \texttt{num\_below} or whose occurrence count per training document is greater than \texttt{fr\_abv}, where the values of \texttt{num\_below} and \texttt{fr\_abv} are taken from Table \ref{table:preprocess-params}. For YRP, we follow the W-LDA paper \cite{nan-etal-2019-topic} and restrict its \texttt{vocab} to only contain top $20,000$ most occurring tokens.

\item Steps 20 through 24 construct the token-to-index map \texttt{w2idx} by mapping each token in \texttt{vocab} to an index starting from 1. Next, we map the padding token to index $0$ (Step 25).

\item The final step in the preprocessing is to encode the train and test documents by mapping each of their tokens to corresponding indices according to \texttt{w2idx}. This is done by the \textsc{encode} function of Algorithm \ref{algorithm:encode-prune} which is invoked in steps 26 and 27.

\end{itemize}

\begin{table}[h]
 \centering
 \resizebox{220pt}{!}{%
 \begin{tabular}{|c|c|c|c|}

 \hline 
  Dataset & \texttt{tr\_size} & \texttt{num\_below} & \texttt{fr\_abv} \\
  \hline 
  AGNews & 96000 & 3 & 0.7 \\
  \hline
  YRP & 448000 & 20 & 0.7 \\
  \hline
\end{tabular}}
\caption{Parameters used for preprocessing the AGNews and YRP datasets.}
\label{table:preprocess-params}
\end{table}

\clearpage


\begin{algorithm}
  \caption{{Pseudocode for preprocessing AGNews and YRP datasets.} 
    \label{algorithm:preprocess}}
  \begin{algorithmic}[1]
    \Function{preprocess}{\texttt{train}, \texttt{test}}
    \Let{\texttt{train}}{\texttt{train}.$sample$(\texttt{tr\_size})}
    \Let{\texttt{tr\_pruned}}{[]}
    \Comment{empty list}
    \Let{\texttt{te\_pruned}}{[]}
    \Comment{empty list}
    \Statex
    \For{document $d$ \textbf{in} \texttt{train}}
        \State \texttt{tr\_pruned}.$append$(\textsc{prune\_doc}($d$))
    \EndFor
    \Statex
    \For{document $d$ \textbf{in} \texttt{test}}
        \State \texttt{te\_pruned}.$append$(\textsc{prune\_doc}($d$))
    \EndFor
    \Statex
    \Let{\texttt{vocab}}{mapping of each token to $0$}
    \Let{\texttt{num\_doc}}{$len$(\texttt{tr\_pruned})}
    
    \Statex
    \For{document $d$ \textbf{in} \texttt{tr\_pruned}}
     \For{token $t$ \textbf{in} $d$}
         \If{$t \notin stopwords$ \textbf{and}\\  
         \pushcode[3] $len(t) \in [3,15]$}
         \State \texttt{vocab}[$t$]$ \gets$ \texttt{vocab}[$t$] $+ 1$
         \EndIf
     \EndFor
 \EndFor
 
    \For{token $t$ \textbf{in} \texttt{vocab}}
         \If{\texttt{vocab}[$t$] $<$ \texttt{num\_below} \textbf{or}\\
         \pushcode[2]\texttt{vocab}[$t$]$/$\texttt{num\_doc} $>$ \texttt{fr\_abv}}
         \State \texttt{vocab}[$t$]$.remove(t)$
         \EndIf
     \EndFor
     \Statex
     \Let{$i$}{1}
     \Let{\texttt{w2idx}}{empty map}
     \For{token $t$ \textbf{in} \texttt{vocab}}
        \State \texttt{w2idx}[$t$]$ = i$
        \State $i \gets i+1$
     \EndFor
     \State \texttt{w2idx}[$0$]$ \gets \mathbb{PAD}$ 
     \Statex
     
     \Let{\texttt{trD}}{\textsc{encode}(\texttt{tr\_pruned}, \texttt{w2idx})}
     \Let{\texttt{teD}}{\textsc{encode}(\texttt{te\_pruned}, \texttt{w2idx})}
     
    \State \Return{\texttt{trD}, \texttt{teD}, \texttt{w2idx}}
    \EndFunction
    \Statex
    
  \end{algorithmic}
\end{algorithm}

\subsection{Learning Rate Scheduler}
As mentioned in section \ref{section:impl-and-train}, we use a learning rate scheduler while training T-TAN. The rate decay follows the following equation:
$$lrate = init\_rate*decay\_rate^{\left \lfloor{\frac{train\_step}{decay\_steps}}\right \rfloor}$$

This is an exponential staircase function which enables decrease in learning rate every epoch during training.\par

\begin{algorithm}
  \caption{{Pseudocode for pruning the document and encoding it given a token-to-index mapping.} 
    \label{algorithm:encode-prune}}
  \begin{algorithmic}[1]
  
  \Function{prune\_doc}{\texttt{doc}}
    \Let{\texttt{doc}}{$rm\_control$(\texttt{doc})}
    \Let{\texttt{doc}}{$rm\_numeric$(\texttt{doc})}
    \Let{\texttt{doc}}{$lowercase$(\texttt{doc})}
    \Let{\texttt{doc}}{$lemmatize$(\texttt{doc})}
    \Let{\texttt{doc}}{$rm\_punctuations$(\texttt{doc})}
    \Let{\texttt{doc}}{$rm\_non\_ASCII$(\texttt{doc})}
    \State \Return{\texttt{doc}}
    \EndFunction
    \Statex
  
  \Function{encode}{\texttt{doc\_list}, \texttt{w2idx}}
    \Let{\texttt{encDocList}}{[]}
    \For{document $d$ \textbf{in} \texttt{doc\_list}}
     \State \texttt{ecDoc }$\gets $ []
         \For{token $t$ in $d$}
          \State \texttt{ecDoc}$.append$(\texttt{w2idx}[$t$])
         \EndFor
    \State \texttt{encDocList}$.append$(\texttt{ecDoc})
     \EndFor
    \State \Return{\texttt{encDocList}}
    \EndFunction
  
  \end{algorithmic}
\end{algorithm}

 We initialize the learning rate by $init\_rate = 0.002$ and use $decay\_rate = 0.96.$ $train\_step$ is a global counter of training steps and $decay\_steps = \frac{\#train\_docs}{batch\_size}$ is the number of training steps taken per epoch. Therefore, effectively, the rate remains constant for all training steps in an epoch and decreases exponentially as per the above equation once the epoch completes.




\subsection{Regularization}
We employ two types of regularization during training:
\begin{itemize}
\item \textbf{Dropout}: We apply dropout \cite{Srivastava2014DropoutAS} to $z$ with the rate of $P_{drop} = 0.6$ before it is processed by the decoder for reconstruction.
\item \textbf{Batch Normalization (BN)}: We apply a BN \cite{Ioffe2015BatchNA} to the inputs of decoder layer and to the inputs of layers being trained for $z_{\mu}$  \& $z_{\log \sigma^2}$, with $\epsilon = 0.001$ and $decay = 0.999$. 
\end{itemize}



\section{Evaluation Metrics}\label{appendix:eval-metric}
Topic models have been evaluated using various metrics namely perplexity, topic coherence, topic uniqueness etc. However, due to the absence of a gold standard for the unsupervised task of topic modeling, all of that metrics have received criticism by the community. Therefore, a consensus on the best metric has not been reached so far. Perplexity has been found to be negatively correlated to topic quality and human judgements \cite{NIPS2009_f92586a2}. This work presents experimental results which show that in some cases models with higher perplexity were preferred by human subjects.\par
Topic Uniqueness \cite{nan-etal-2019-topic} quantifies the intersection among topic words globally. However, it also suffers from drawbacks and often penalizes a model incorrectly \cite{hoyle-etal-2020-improving}. Firstly, it  does not account for ranking of intersected words in the topics. Secondly, it fails to distinguish between the following two scenarios: \textbf{1)} When the intersected words in one topic are all present in a second topic (signifying strong similarity i.e. these two topics are essentially identical) and,  \textbf{2)} When the intersected words of one topic are spread across all the other topics (signifying weak similarity i.e. the topics are diffused). The first is a problem related to uniqueness among topics while second is a problem related to word intrusion in topics. \cite{NIPS2009_f92586a2} conducted experiments with human subjects on two tasks: word intrusion and topic intrusion. Word intrusion measures the presence of those words (called intruder words) which disagree with the semantics of the topic. Topic intrusion measures the presence of those topics (called intruder topics) which do not represent the document corpus appropriately. These are better estimates of human judgement of topic models in comparison to perplexity and uniqueness. However, since these metrics rely on human feedback, they cannot be widely used for unsupervised evaluation. Further, topic uniqueness unfairly penalizes cases when some words are common between topics, however other uncommon words in those topics change the context as well as topic semantics as also discussed in \cite{hoyle-etal-2020-improving}.
\noindent According to the work of \cite{lau-etal-2014-machine}, measuring the normalized pointwise mutual information (NPMI) between all the word pairs in a set of topics agrees with human judgements most closely. This is called the NPMI Topic Coherence in the literature and is widely used for the evaluation of topic models. We therefore adopt this metric in our work. Since the effectiveness of a topic model actually depends on the topic representations that it extracts from the documents, we report the performance of our model on two downstream tasks: document classification and keyphrase generation (which use these topic representations) for a better and holistic evaluation and comparison.

\begin{table}[h!]
 \centering
 \begin{tabular}{|p{7cm}|}
 \hline 
 \textbf{Would a pilot know that one of their crew is armed?}\\
 The Federal Flight Deck Officer page on Wikipedia says this:\\ \\
 Under the FFDO program, flight crew members are authorized to use firearms. A flight crew member may be a pilot, flight engineer or navigator assigned to the flight.\\ \\
 To me, it seems like this would be crucial information for the PIC to know, if their flight engineer (for example) was armed; but on the flip-side of this, the engineer might want to keep that to himself if he's with a crew he hasn't flown with before.\\ \\
 Is there a guideline on whether an FFDO should inform the crew that he's armed?\\ \\
 \textbf{GT:} \textit{security, crew, ffdo}\\
 \textbf{TAKG:} \textit{faa regulations, ffdo, flight training, firearms, far}\\
 \textbf{TAKG + W-TAN:} \textit{ffdo, crew, flight controls, crewed spaceflight, security}\\
  \hline
  \textbf{Do the poisons in “Ode on Melancholy” have deeper meaning?}\\
  In "Ode on Melancholy", Keats uses the images of three poisons in the first stanza: Wolf's bane, nightshade, and yew-berries. Are these poisons simply meant to connote death/suicide, or might they have a deeper purpose?\\ \\
  \textbf{GT:} \textit{poetry, meaning, john keats}\\
 \textbf{TAKG:} \textit{the keats, meaning, poetry, ode, melancholy keats}\\
 \textbf{TAKG + W-TAN:} \textit{poetry, meaning, the keats, john keats, greek literature}\\
  \hline
\end{tabular}
\caption{Two randomly selected posts (title in \textbf{bold}) from StackExchange dataset with ground truth (\textbf{GT}) and top 5 keyphrases predicted by TAKG with and without W-TAN, denoted as \textbf{TAKG + W-TAN} \&  \textbf{TAKG} respectively. Keyphrases generated with W-TAN are closer to the ground truth in terms of both prediction and ranking.}
\label{table:qual-TAKG-KP}
\end{table}

\section{Qualitative Analysis}
\subsection{Key Phrase Predictions}
We saw the quantitative improvement in results in Table \ref{table:takg} when we used W-TAN as the topic model with TAKG. In Table \ref{table:qual-TAKG-KP}, we display some posts from StackExchange dataset with ground truth keyphrases and top 5 predictions by TAKG with and without W-TAN. We observe that using W-TAN improves keyphrase generation qualitatively.\par
The first post in Table \ref{table:qual-TAKG-KP} inquires if a flight officer should inform the pilot in command (PIC) about him being armed or not. For this post, TAKG alone only predicts one ground truth keyphrase correctly and misses \textit{\lq security\rq} and \textit{\lq crew\rq}. However, when TAKG is used with W-TAN, it gets all three ground truth keyphrases, two of which are its top 2 predictions as well.\par
The second post is inquiring about a possible deeper meaning of three poisons in a poem by John Keats. TAKG alone predicts two of the ground truth keyphrases correctly but assigns them larger ranks and it misses \textit{\lq john keats\rq}.
When TAKG is used with W-TAN, it gets all three ground truth keyphrases and its top 2 keyphrases are assigned the exact same rank as they have in the ground truth. This hints that using W-TAN with TAKG improves the prediction as well as ranking of the generated keyphrases compared to using TAKG alone.

\end{document}